\documentclass[letterpaper]{article} % DO NOT CHANGE THIS
\usepackage{aaai2026}  % DO NOT CHANGE THIS
\usepackage{times}  % DO NOT CHANGE THIS
\usepackage{helvet}  % DO NOT CHANGE THIS
\usepackage{courier}  % DO NOT CHANGE THIS
\usepackage[hyphens]{url}  % DO NOT CHANGE THIS
\usepackage{graphicx} % DO NOT CHANGE THIS
\usepackage{natbib}  % DO NOT CHANGE THIS AND DO NOT ADD ANY OPTIONS TO IT
\usepackage{caption} % DO NOT CHANGE THIS AND DO NOT ADD ANY OPTIONS TO IT
\usepackage{algorithm}
\usepackage{algorithmic}

\definecolor{headerblue}{RGB}{0,91,119}
\definecolor{lightgray}{gray}{0.9}
\usepackage{colortbl}
\usepackage{multirow}
\usepackage{booktabs}
\usepackage{subcaption}
\usepackage[most]{tcolorbox}
\usepackage{makecell}

\usepackage{mathtools}
\usepackage{amssymb}
\usepackage{amsthm}
\usepackage{color}
\newcommand{\cut}[1]{{}}
\newcommand{\vh}{{\mathbf{h}}}

\newcommand{\vs}{{\mathbf{s}}}

\newcommand{\vx}{{\mathbf{x}}}
\newcommand{\vy}{{\mathbf{y}}}

\newcommand{\vK}{{\mathbf{K}}}

\newcommand{\vM}{{\mathbf{M}}}

\newcommand{\vP}{{\mathbf{P}}}
\newcommand{\vQ}{{\mathbf{Q}}}

\newcommand{\vV}{{\mathbf{V}}}
\newcommand{\vW}{{\mathbf{W}}}
\newcommand{\vX}{{\mathbf{X}}}

\newcommand{\vZ}{{\mathbf{Z}}}

\newcommand{\cB}{{\mathcal{B}}}
\newcommand{\cC}{{\mathcal{C}}}

\newcommand{\cI}{{\mathcal{I}}}

\newcommand{\cM}{{\mathcal{M}}}

\newcommand{\cY}{{\mathcal{Y}}}

\makeatletter
\let\@@span\span
\def\sp@n{\@@span\omit\advance\@multicnt\m@ne}
\makeatother

\newcommand{\bc}{\begin{center}}
\newcommand{\ec}{\end{center}}

\newcommand{\bdm}{\begin{displaymath}}
\newcommand{\edm}{\end{displaymath}}

\newcommand{\beq}{\begin{equation}}
\newcommand{\eeq}{\end{equation}}

\newcommand{\bfl}{\begin{flushleft}}
\newcommand{\efl}{\end{flushleft}}

\newcommand{\bt}{\begin{tabbing}}
\newcommand{\et}{\end{tabbing}}

\newcommand{\beqn}{\begin{align}}
\newcommand{\eeqn}{\end{align}}

\newcommand{\beqs}{\begin{align*}} % no equation numbers
\newcommand{\eeqs}{\end{align*}}  % no equation numbers

\newtheorem{remark}{Remark}

\usepackage{newfloat}
\usepackage{listings}
\DeclareCaptionStyle{ruled}{labelfont=normalfont,labelsep=colon,strut=off} 
\floatstyle{ruled}
\newfloat{listing}{tb}{lst}{}
\floatname{listing}{Listing}
\title{Decoding Memories: An Efficient Pipeline for \\ Self-Consistency Hallucination Detection 
}
\author{
    Weizhi Gao\textsuperscript{1}\thanks{This work was done during the internship at ORNL}, 
    Xiaorui Liu\textsuperscript{1},
    Feiyi Wang\textsuperscript{2},
    Dan Lu\textsuperscript{2},
    Junqi Yin\textsuperscript{2}
}
\affiliations{
    \textsuperscript{\rm 1}Department of Computer Science, North Carolina State University \\
    \textsuperscript{\rm 2}Oak Ridge National Laboratory
}

\usepackage{bibentry}

\begin{document}

\nocopyright
\maketitle
\footnotetext[1]{Preprint. Under review. This version: \today.}

\begin{abstract}
Large language models (LLMs) have demonstrated impressive performance in both research and real-world applications, but they still struggle with hallucination. Existing hallucination detection methods often perform poorly on sentence-level generation or rely heavily on domain-specific knowledge. While self-consistency approaches help address these limitations, they incur high computational costs due to repeated generation. In this paper, we conduct the first study on identifying redundancy in self-consistency methods, manifested as shared prefix tokens across generations, and observe that non-exact-answer tokens contribute minimally to the semantic content. Based on these insights, we propose a novel Decoding Memory Pipeline (DMP) that accelerates generation through selective inference and annealed decoding. Being orthogonal to the model, dataset, decoding strategy, and self-consistency baseline, our DMP consistently improves the efficiency of multi-response generation and holds promise for extension to alignment and reasoning tasks. Extensive experiments show that our method achieves up to a 3x speedup without sacrificing AUROC performance.
\end{abstract}

\begin{links}
    \link{Code}{https://anonymous.4open.science/status/Decoding-Memorie-Pipeline-4E8F}
\end{links}

\section{Introduction}
\label{sec:introduction}

Large Language Models (LLMs) have made significant advancements in both research and practical applications~\citep{openai2023gpt, llama3modelcard}. However, a persistent challenge is their tendency to hallucinate, where the model produces outputs that are grammatically correct and fluent, yet factually inaccurate, especially in sentence-level generation~\citep{park2025steer, ji2023survey}. The difficulty of identifying factual error poses a substantial barrier to deploying LLMs in reliability-critical domains such as scientific research and medical diagnostics~\citep{athaluri2023exploring, pal2023med, chai2024exploring}. Therefore, it is urgent to design methods that accurately detect hallucination and reject unreliable answers~\citep{manakul2023selfcheckgpt, chen2024inside}.

Recent research has explored various approaches for detecting hallucinations in LLM outputs. One line of work estimates the likelihood of generated content using metrics such as perplexity and energy scores~\citep{ren2022out, liu2020energy}. However, the scores using likelihood are affected by perturbations in grammar and common word usage, making them less effective at accurately identifying hallucinations at the sentence level~\citep{chen2024inside}. Another approach involves linear probing of hidden states~\citep{azaria2023internal, li2025hd}, nevertheless, requiring labeled training data and often struggling to generalize to out-of-distribution examples~\citep{pan2023fact, manakul2023selfcheckgpt}. Moreover, due to the overconfident nature of LLMs, these methods fail to provide reliable confidence estimates for hallucination detection~\citep{xiong2023can}.

Given the limitations of existing approaches, self-consistency methods have garnered significant attention for hallucination detection~\citep{farquhar2024detecting, chen2024inside}. These methods assess the consistency among multiple responses to the same query, using metrics such as semantic entropy. They not only achieve strong performance in detecting hallucinations but also exhibit robustness across diverse domains~\citep{manakul2023selfcheckgpt}. Moreover, the ensemble nature enables uncertainty quantification, allowing them to produce confidence scores for hallucination detection. However, a key drawback of self-consistency methods is their high computational cost, as they require generating multiple responses per query. It often incurs 5 to 10 times the cost, making them less suitable for resource-constrained applications~\citep{farquhar2024detecting}.

In this work, we present the first preliminary study that investigates the redundancy inherent in self-consistency methods based on multiple generations. We observe that, for a broad range of questions, the generated multiple responses often share similar tokens and prefixes, indicating overlap in token sequences. Furthermore, non-exact-answer tokens that do not contribute substantively to the core content of the answer, such as common function words ``a" and ``the", have minimal impact on consistency evaluation. Therefore, the redundancy is therein though  responses share different template caused by those tokens. We identify that two kinds of redundancy are widely present across models and datasets, constituting a substantial portion of the generated content.

To exploit the redundancy in multiple generations, we propose a Decoding Memory Pipeline (DMP) to accelerate self-consistency methods. Specifically, we introduce selective inference to leverage the autoregressive pattern of language models, which skips redundant forward passes by reusing cached computations from overlapped prefixes across generations. To further increase the portion of computation that can be skipped, we propose annealed decoding, which progressively lowers the sampling temperature to make non-exact-answer tokens more deterministic during generation. Our DMP is agnostic to self-consistency methods, enabling broad applicability for hallucination detection.
To evaluate the effectiveness of our proposed DMP, we conduct extensive experiments on several benchmark datasets, including TriviaQA, NQ-Open, SQuAD, and HaluEval using a range of strong baselines. Our results show that DMP adequately reduces the redundancy, and accelerate the generation process by up to 3x without compromising hallucination detection performance.
In summary, our main contributions are as follows:
\begin{itemize}
    \item To the best of our knowledge, we are the first to identify redundancy in self-consistency methods in hallucination detection. Specifically, we reveal that redundancy arises from shared prefix tokens and non-exact-answer tokens across models and datasets.
    \item We propose a novel Decoding Memory Pipeline (DMP) that is agnostic to self-consistency methods, enabling accelerated generation. Specifically, we introduce selective inference, which skips redundant computations by leveraging repeated tokens, and annealed decoding, which increases token reuse by mitigating the impact of non-exact-answer tokens during generation.
    \item Extensive experiments on TriviaQA, NQ-Open, SQuAD, and HaluEval with multiple self-consistency methods demonstrate that our DMP accelerates generation by up to 3x without compromising hallucination detection performance across multiple self-consistency baselines.
\end{itemize}

\section{Related Works}
\label{sec:related_works}

\subsection{Efficient LLMs}
LLMs face significant challenges in efficient deployment, prompting extensive research into improving inference efficiency~\citep{wan2023efficient}. To address the quadratic complexity of the attention mechanism, key-value (KV) cache has become a standard inference strategy, enabling reuse of past key-value pairs~\citep{hooper2024kvquant, liu2024minicache}. Additionally, attention masks are employed to facilitate batch-wise inference~\citep{vaswani2017attention, devlin2019bert}. Various approaches, such as pruning~\citep{ma2023llm}, quantization~\citep{gao2025modulated, lin2024awq}, and knowledge distillation~\citep{xu2024survey}, have been proposed to further accelerate inference. However, these methods are generally not tailored to self-consistency techniques, which incur computational redundancy due to generating multiple responses for the same input prompt. Our DMP is compatible with both KV caching and batch-wise inference, and is orthogonal to existing inference acceleration techniques.

\subsection{Hallucination Detection}
Hallucinations in LLMs present major challenges for reliable deployment, leading to growing interest in effective detection methods~\citep{ji2023survey, rawte2023survey}. One common approach estimates hallucination scores by computing generation likelihoods, using metrics such as perplexity or energy scores~\citep{ren2022out, liu2020energy}. However, these methods often fail to achieve satisfactory detection performance. Other works attempt to identify hallucinations by training classifiers on hidden states~\citep{azaria2023internal, li2025hd, su2024unsupervised, orgad2024llms}, but these approaches require labeled data and struggle with out-of-domain generalization~\citep{pan2023fact, manakul2023selfcheckgpt}. Furthermore, the overconfidence of LLMs limits the reliability of such confidence-based detection~\citep{xue2023repeat}. Self-consistency methods, such as lexical similarities, SelfCheckGPT, and semantic entropy, offer an alternative by evaluating consistency across multiple responses to the same query~\citep{manakul2023selfcheckgpt, farquhar2024detecting, chen2024inside, lin2022towards}. Nevertheless, these methods are hindered by their high computational cost.

\section{Background}
\label{sec:background}
In this section, we provide essential background on LLM generation and self-consistency methods.
\begin{figure*}[!ht]
    \centering
    \includegraphics[width=0.7\linewidth]{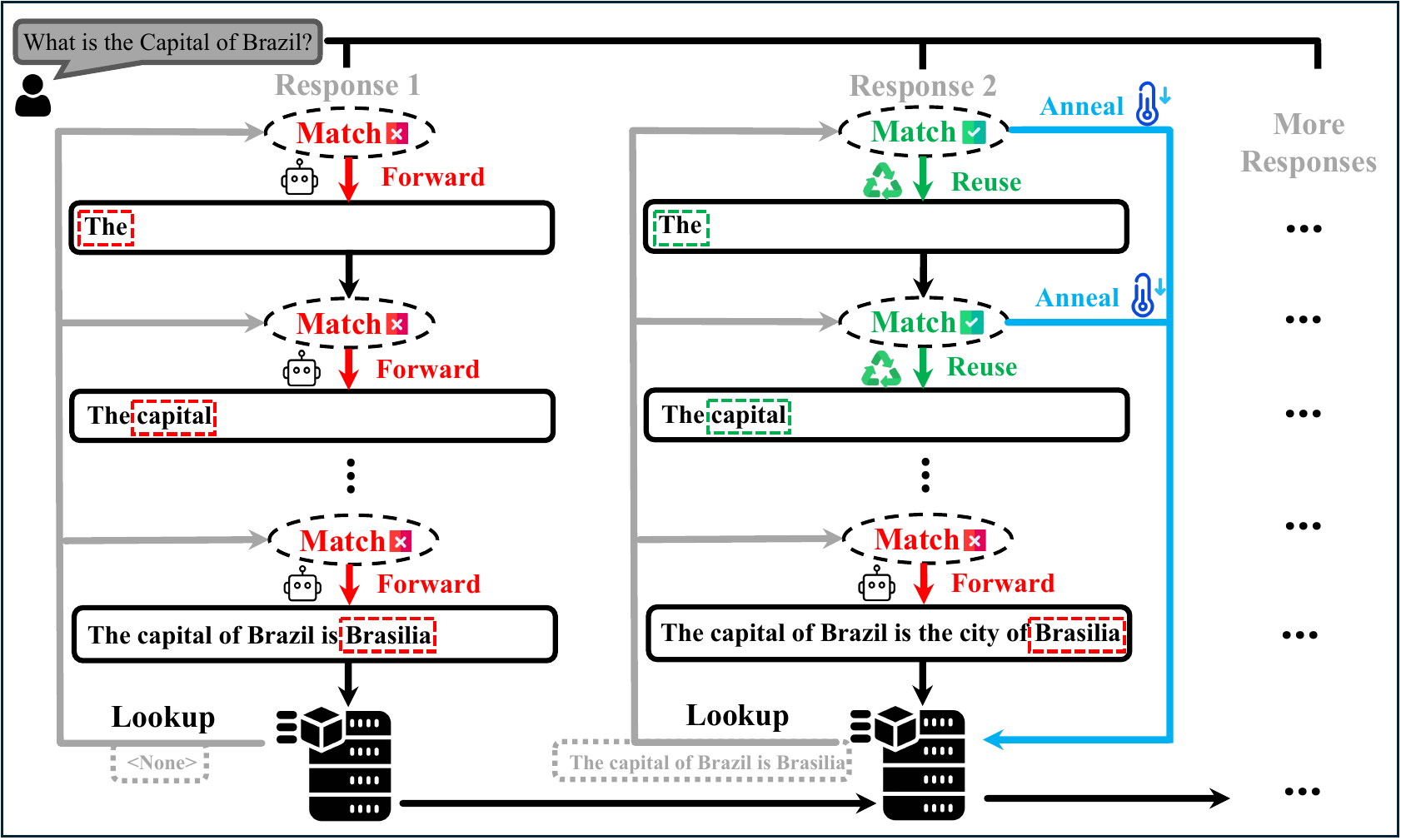}
    \caption{The illustration of our DMP. Given the input prompt, the model generates multiple responses. The current generation is compared with cached results: matches are reused as tokens ({Green}), while mismatches trigger a forward pass through the LLM ({Red}). Annealing ({Blue}) is used to increase the reuse ratio by lowering the cached sampling temperature for less important tokens. After generating a complete response, we add it to the cache if it is not already present.}
    \label{fig:DMP}
\end{figure*}

\subsection{LLM Generation Pipeline}
LLMs typically adopt an autoregressive generation paradigm, which involves alternating between inference and decoding steps~\citep{henighan2020scaling}. In the inference stage, the model computes a logit for the next token, conditioned on the input context and all previously tokens. Given the input prompt $\vx$, the logit of generating the $i$-th token $\vy_i$ is computed as follows:
\begin{align}
\label{eqn:llm_forward}
\vs_i = f_\theta(\vx, \vy_{1:i-1}), \quad i = 1, 2, \ldots, L,
\end{align}
where $f_\theta$ denotes a transformer-based model parameterized by $\theta$, and $L$ is the total number of tokens to be generated. In the decoding stage, a multinomial distribution over the token $\vy_i$ is constructed based on its logit. Given a sampling temperature $T$, the probability is computed as:
\begin{align}
\label{eqn:llm_prob}
\vP(\vy_i \mid \vx, \vy_{1:i-1}) = \text{softmax}\left(\frac{\vs_i}{T}\right), i = 1, 2, \ldots, L,
\end{align}
where the next token $\vy_i$ is sampled from this distribution. The process repeats until an end-of-sequence token is produced. To trade off between diversity and determinism, various sampling strategies are employed, including greedy decoding, beam search, top-$k$, and top-$p$ sampling. 

\textbf{KV Cache.} To reduce the computational burden of the attention mechanism, past key $\vK_{1:i-1}$ and value tensors $\vV_{1:i-1}$ are cached to avoid redundant computation. The forward process with KV cache is updated as:
\begin{align}
    \vs_i = f_\theta(\vx, y_{i-1}, \vK_{1:i-2}, \vV_{1:i-2}), \quad i = 2, 3, \ldots, L.
\end{align}
Despite being optimized by KV cache, the sequential nature of generation remains computationally intensive, as it requires $O(L)$ model forward passes for a sequence of length $L$. For more details, please refer to Appendix A.

\subsection{Self-Consistency Hallucination Detection}
Despite their impressive performance across a wide range of applications, LLMs are prone to hallucinations, where the responses appear plausible but contain factual inaccuracies. To address this issue, self-consistency methods have been proposed, which not only detect hallucination without training but also provide uncertainty of the prediction. Given an input prompt $\vx$, suppose we generate $N$ responses to evaluate consistency, denoted as a set $\cY = \{\vy_1, \vy_2, \dots, \vy_N\}$. A consistency score $S(\cY; \vx)$ is then computed to quantify the agreement among these responses. For example, the Length-Normalized Entropy evaluates the normalized likelihood of the generated responses~\citep{malinin2020uncertainty}:
\begin{align}
    S_\text{LNE}(\cY|\vx) = -\mathbb{E}_{\vy\in\cY}\frac{1}{L_\vy}\sum_{i=1}^{L_\vy}\log\vP(\vy_i \mid \vx, \vy_{1:i-1}), 
\end{align}
where a higher consistency score means a more factual answer. However, generating multiple responses further increases the computational cost of hallucination detection.

\section{Decoding Memory Pipeline}
\label{sec:method}
In this section, we introduce our decoding memory pipeline. We begin by presenting preliminary studies that reveal the computational redundancy in self-consistency generation. To address this, we propose a selective inference mechanism that reduces computational cost by selectively skipping forward passes using cached logits in Section~\ref{subsec:selective_inference}. We further introduce annealed decoding, a technique designed to enhance token reuse by progressively lowering the sampling temperature for non-exact-answer tokens in Section~\ref{subsec:annealed_decoding}. An overview of our proposed pipeline is provided in Figure~\ref{fig:DMP}.

\subsection{Redundancy in Self-Consistency Generation}
\label{subsec:preliminary}
We conduct preliminary studies to investigate computational redundancy in self-consistency hallucination detection. Specifically, we make the following two observations when generating multiple responses.

\begin{tcolorbox}[colback=gray!20, colframe=black]
\small
\label{obs1}
Observation 1: Repeated Prefix in Multiple Responses.
\end{tcolorbox}

\begin{figure}[!t]
    \centering
    \begin{subfigure}{0.49\linewidth}
        \centering
        \includegraphics[width=\linewidth]{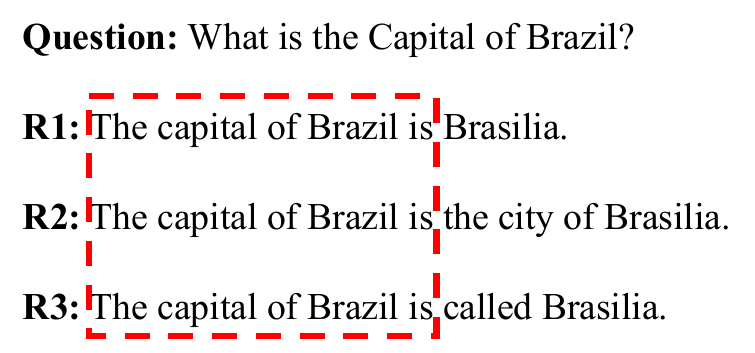}
        \caption{}
        \label{fig:prelim_prefix_example}
    \end{subfigure}
    \hfill
    \begin{subfigure}{0.49\linewidth}
        \centering
        \includegraphics[width=\linewidth]{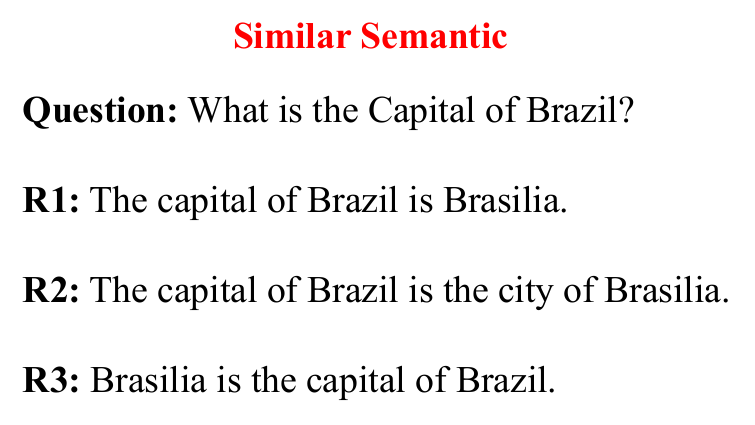}
        \caption{}
        \label{fig:prelim_noexact_example}
    \end{subfigure}

    \begin{subfigure}{0.49\linewidth}
        \centering
        \includegraphics[width=\linewidth]{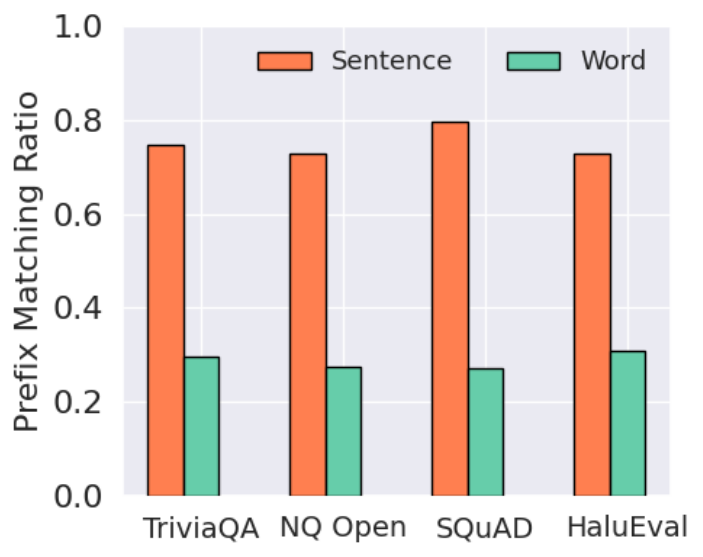}
        \caption{}
        \label{fig:prelim_prefix}
    \end{subfigure}
    \hfill
    \begin{subfigure}{0.49\linewidth}
        \centering
        \includegraphics[width=\linewidth]{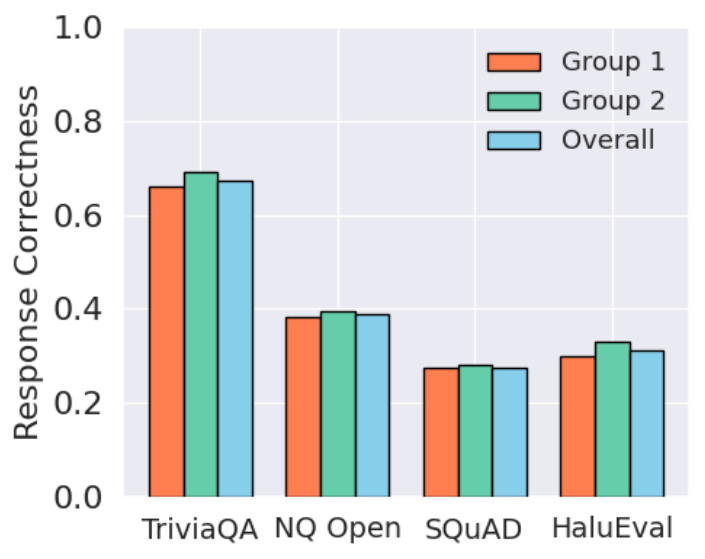}
        \caption{}
        \label{fig:prelim_noexact}
    \end{subfigure}
    
    \caption{Preliminary study of redundancy with Llama2-7B-Chat. (a) Example of shared prefixes across multiple generations. (b) Example that variations in non-exact-answer tokens preserve the semantics. (c) Prefix matching ratios at the sentence level and word level across datasets. (d) Response accuracy after template-based grouping across datasets.}
    \label{fig:prefix}
\end{figure}

Intuitively, responses generated for the same input prompt often share overlapping content, which is dependent on the model confidence. Since we are working with autoregressive models and aim to exploit this redundancy, we focus on the repetition of prefixes, i.e., the initial segments of the responses. For example, given the question \textit{``What is the capital of Brazil?''}, multiple responses commonly share the prefix \textit{``The capital of Brazil is''} as shown in Figure~\ref{fig:prelim_prefix_example}.

To verify the presence of redundancy, we analyze prefix overlap in multiple-response generation using Llama2-7B-Chat~\cite{touvron2023llama} across several datasets. We evaluate overlap from two perspectives: (1) the proportion of response pairs that share prefixes, and (2) the proportion of words in the shared prefixes relative to the total word count. As shown in Figure~\ref{fig:prefix}, the sentence-level and the word-level ratio exceeds $70\%$ and $25\%$ across all four datasets, respectively. These results demonstrate that redundancy is prevalent across both datasets and models. For additional implementation details and results on other models, please refer to Appendix B.1 and D.1.

\begin{tcolorbox}[colback=gray!20, colframe=black]
\small
\label{obs2}
Observation 2: Non-exact-answer tokens do not matter.
\end{tcolorbox}

Recent research has shown that extracting exact answers is critical for improving performance in many tasks~\citep{orgad2024llms, asai2020logic}, as they capture the key information within responses. Consequently, hallucination detection should primarily rely on the exact answers, while non-exact-answer tokens affect sentence structure or word order. As shown in Figure~\ref{fig:prelim_noexact_example}, given the question \textit{What is the capital of Brazil?}, the non-exact-answer tokens vary, but the responses convey the same semantic information, indicating the inner redundancy.

To test the generality of this assumption, we conduct experiments to assess the influence of non-exact-answer tokens. Specifically, for each question, we generate multiple responses and then group them based on their answer templates using GPT-3.5~\citep{openai2023gpt}. The rationale is that non-exact-answer tokens primarily affect the template or word order of the responses. After clustering the responses, we compare the accuracy of the two groups. As shown in Figure~\ref{fig:prelim_noexact}, the accuracies of different groups are nearly identical to the overall accuracy, a trend consistent across datasets. This suggests that non-exact-answer tokens have a limited impact on the semantics of the generated responses. For implementation details, please refer to Appendix B.1.

\subsection{Selective Inference}
\label{subsec:selective_inference}
To mitigate the redundancy identified in Observation 1, we propose a selective inference strategy that skips unnecessary generation steps. The high-level motivation is that LLM inference will generate the same probability distribution with the same previous tokens, given the autoregressive pattern in Equation~\ref{eqn:llm_forward}. To exploit this, we define a memory list $\cM$ to store the response memory, which is initialized as an empty list. Then we define response memories to cache generated responses, where each entry stores the generated tokens $\vy$ and their associated logits $\vs$: 
\begin{align}
\vM_m = (\mathbf{y}^{(m)}, \mathbf{s}^{(m)}),
\end{align}
where $m$ denotes the index of a cached generated response. During the generation of token $\vy_i^{(n)}$, we iterate over the memory entries $\vM_m\in\cM$ to check whether the prefix $\vy_{1:i-1}^{(n)}$ matches any stored response $\vy_{1:i-1}^{(m)}$. If a match is found, we directly assign the corresponding $\vs_i^{(m)}$ to $\vy_i^{(n)}$, and reuse $\vs_i^{(m)}$ to predict the next token, bypassing the model inference step. If no matching prefix is found, a forward pass is performed: $\vs_i^{(n)} = f_\theta(\vx, \vy_{1:i-1}^{(n)})$. After generating a newly complete response $\vy^{(n)}$, it will be added to the memory list $\cM$ if it is not already present.

\textbf{KV Cache.} Selective inference remains compatible with KV caching by caching the key and value tensors of responses. Specifically, we augment the response memory with keys $\vK^{(m)}$ and values $\vV^{(m)}$ of the whole sequence:
\begin{align}
    \vM_m = (\mathbf{y}^{(m)}, \mathbf{s}^{(m)}, \vK^{(m)}, \vV^{(m)}).
\end{align}
We assign $\vK_i^{(m)}$ and $\vV_i^{(m)}$ to $\vy_i^{(n)}$ if a matching prefix is found in the memory set $\cM$. If no match is found, a forward pass is performed using the KV cache:
$\vs_i^{(n)} = f_\theta(\vx, \vy_{i-1}^{(n)}, \vK_{1:i-2}^{(n)}, \vV_{1:i-2}^{(n)})$.
Note that the outputs produced by our selective inference method are equivalent to those generated by the standard generation pipeline. 

\textbf{Batch Mode.} Our DMP supports batch-wise inference, aligning with the parallel processing characteristics of deep learning. This is achieved through a reusing mask, where the model continues its forward pass until no prefix match can be identified under this mask. To be specific, the reusing mask $\mu_r$ is initialized as the attention mask in causal language models~\citep{vaswani2017attention}, where only the input prompt tokens are unmasked. Let the generated batch be $B_G$, and a reference batch $B_R$ of the same size be initialized with all entries set to \textit{None}. At each generation step, we first perform a batch-wise prefix match between $B_G$ and $B_R$ under the attention mask $\mu_a$. For $j$-th instance in the batch, the reusing mask $\mu_r$ is updated as follows:
\begin{itemize}
    \item For instance $B_G[j]$ that fails to match, set all the items of $\mu_r[j]$ as $ 0$ and update $B_R[j]$ with $\textit{Next}(\cM_{B_G[j]})$.
    \item For instance $B_G[j]$ that matches, $\mu_r[j]$ is updated to follow $\mu_a[j]$, with the next token position unmasked.
\end{itemize}

Here, $\cM_{B_G[j]}$ is the memory list of $B_G[j]$, and \textit{Next}$(\cM)$ fetches the next cached response in $\cM$. For the instances with a successful match, both the attention and generated batches are updated as follows:
\begin{align}
\mu_r(B_G) &= \mathrm{Update}(\mu_r(B_R)),\\
\mu_r(\mu_a) &= \mu_r,
\end{align}
where $\mathrm{Update}(\cdot)$ denotes the next token information update in selective inference. If no match is found after iterating through all the caches in $\cM_{B_G}$, we perform a forward process over the batch $f(B_G, \mu_a)$. Each newly generated complete response $B_G[j]$ is added to the memory list $\cM_{B_G[j]}$ if it is not already present. The complete batch-wise inference process is outlined in Appendix C, where all steps are implemented in a fully parallelization manner.

\subsection{Annealed Decoding}
\label{subsec:annealed_decoding}
As indicated in Observation 2, non-exact tokens have minimal influence on semantics and hallucination detection performance. To enhance template consistency in responses, we introduce annealed decoding to mitigate the impact of non-exact-answer tokens. To implement this, we first identify less important tokens using the cosine similarity. Then, during decoding, we progressively lower the sampling temperature for these tokens based on how frequently they appear, effectively promoting the reuse ratio across generations.

To identify less important tokens, we compute the cosine similarity between each response token embedding and the embedding of the input prompt. Intuitively, tokens that exhibit high similarity to the prompt embedding are likely to be semantically entailed or repeated from the prompt itself, and thus contribute less new information. In contrast, tokens with low similarity are more likely to introduce novel content not directly present in the prompt, making them informative and important. Therefore, we treat high-similarity tokens as non-exact-answer tokens and target them for temperature annealing. First, we augment the response memory $\vM_m$ by caching the hidden states $\vh^{(m)} = [\vh(\vx), \vh(\vy^{(m)})]$, where $\vh(\cdot)$ denotes the final-layer hidden embeddings. The importance score of a response token $\vy_i^{(k)}$ is computed as:
\begin{align}
    I_i^{(m)} = -\cos{\left(\vh(\vx_{-1}), \vh(\vy_i^{(m)})\right)},
\end{align}
where we represent the input prompt using the embedding of its last token, $\vh(\vx_{-1})$. Then we select less important tokens by comparing them with the average score:
\begin{align}
    \cI^{(m)} = \{i \mid I_i^{(m)}<\frac{\alpha}{L^{(m)}}\sum_{i=1}^{L^{(m)}}I_i^{(m)}\},
\end{align}
where the hyperparameter $\alpha$ controls the selection threshold. Based on the occurrence frequency of non-exact-answer tokens, we reduce the sampling randomness of them, as more frequent tokens tend to form a preferable template for generation. In particular, when selective inference skips generation using the $m$-th cached response and the $i$-th token is identified as a non-exact-answer token, we scale up and update the cached logits $\vs^{(m)}$ as follows:
\begin{align}
\vs_i^{(m)} = \eta * \vs_i^{(m)} , \quad \text{if } i \in \cI^{(m)},
\end{align}
where $\eta \in (1, \infty)$ is the annealing speed for controlling the strength of annealing. Our technique is equivalent to lowering the sampling temperature and encourages deterministic reuse of non-exact-answer tokens while preserving the informative parts of the sequence. 

\begin{algorithm}[tb]
\small
\caption{Decoding Memory Pipeline}
\label{alg:algorithm}
\textbf{Input}: Input prompts $\vx$, Number of Responses $N$, model $f$\\
\textbf{Parameter}: Sampling Temperature $T$, Confidence Threshold $\gamma$, Rescaling $\eta$, Selection Threshold $\alpha$\\
\textbf{Output}: $N$ responses
\begin{algorithmic}[1] %[1] enables line numbers
\STATE Initailze memory list $\cM=[\quad]$;
\FOR{$i=1:N$}
\STATE Initialize the current response $\vy_i=\vx$;
\FOR{$j=1:$max\_generation}
\FOR{$k=1:\text{len}(\cM)$}
\IF {$\cM[k]$ start with $\vy_i$}
\STATE Hard Decoding with cache and inherit scores $\vs_i$, KV cache $K_i$, $V_i$, and hidden states $\vh_i$;
\ELSE
\STATE $\vy_i[j+\text{len}(x)]$ = $f(\vx, K_i, V_i)$;
\ENDIF
\ENDFOR
\ENDFOR
\IF {$\vy_i$ is not contained in $\cM$}
\STATE $\cM$.append(($\vy_i$, $\vs_i$, $K_i$, $V_i$, $\vh_i$));
\ELSE
\STATE Anneal the memory in $\cM$ corresponding to $\vy_i$;
\ENDIF
\ENDFOR
\STATE \textbf{return} $[\vy_1, \vy_2, \dots, \vy_N]$
\end{algorithmic}
\end{algorithm}

\textbf{Hard Decoding.} For tokens with highly concentrated sampling distributions during generation, resampling often results in negligible variation and does not alter the semantic outcome. To increase the reuse efficiency in such cases, we introduce hard sampling, which deterministically reuses the token prediction when the model is confident. Specifically, if the maximum sampling probability exceeds a threshold $\gamma$ and the cached token is correspond to the maximum, we sample the next token as follows:
\begin{align*}
\vs_i^{(n)} = \vy_i^{(m)}, \quad
&\text{if } \max(\vP(\vy_i^{(m)} \mid \vx, \vy^{(m)}_{1:i-1})) > \gamma \\
&\text{and } \vy_i^{(m)}=\arg\max\vs_i^{(m)},
\end{align*}
where the hyperparameter $\gamma$ controls the confidence threshold for applying hard decoding.

\begin{table*}[!ht]
\centering
\small
\begin{tabular}{l|cccc|cccc}
\toprule
\multicolumn{1}{c|}{} & 
\multicolumn{4}{c|}{\textbf{Llama2-7B-Chat}} & 
\multicolumn{4}{c}{\textbf{Mistral-7B-Instruct}} \\
\multicolumn{1}{c|}{} & \textbf{TriviaQA} & \textbf{NQ Open} & \textbf{SQuAD} & \textbf{HaluEval} & \textbf{TriviaQA} & \textbf{NQ Open} & \textbf{SQuAD} & \textbf{HaluEval} \\
\midrule
LN-Entopy & 69.5 & 61.2 & 66.9 & 68.2 & 73.2 & 65.7 & 59.9 & 71.2 \\
\rowcolor{lightgray}
LN-Entropy+DMP & 71.0 & 64.4 & 68.6 & 69.8 & 73.3 & 67.8 & 61.0 & 73.8 \\
\midrule
Lexical Similarity  & 66.7 & 62.9 & 70.0 & 61.5 & 75.8 & 68.2 & 64.8 & 79.3 \\
\rowcolor{lightgray}
Lexical Similarity+DMP & 67.7 & 62.0 & 70.8 & 65.3 & 75.5 & 67.8 & 63.9 & 78.0 \\
\midrule
Semantic Entropy & 86.4 & 81.8 & 82.5 & 77.9 & 84.6 & 79.9 & 76.7 & 84.3  \\
\rowcolor{lightgray}
Semantic Entropy+DMP & 85.4 & 82.0 & 81.5 & 77.2 & 84.8 & 78.9 & 77.1 & 84.9 \\
\midrule
Semantic Entropy-B & 86.3 & 81.6 & 82.2 & 78.0 & 84.8 & 80.0 & 77.0 & 84.5  \\
\rowcolor{lightgray}
Semantic Entropy-B+DMP & 86.4 & 82.3 & 81.4 & 77.0 & 84.6 & 79.1 & 77.5 & 84.8 \\
\midrule
SelfCheckGPT & 64.9 & 64.9 & 70.5 & 61.2 & 71.4 & 64.5 & 65.1 & 74.6 \\
\rowcolor{lightgray}
SelfCheckGPT+DMP & 67.5 & 62.9 & 70.3 & 64.5 & 71.6 & 66.9 & 60.1 & 76.0 \\
\midrule
EigenScore & 70.6 & 65.8 & 70.0 & 66.3 & 75.1 & 66.5 & 65.2 & 79.9 \\
\rowcolor{lightgray}
EigenScore+DMP & 69.5 & 63.9 & 71.8 & 68.3 & 75.8 & 67.3 & 66.4 & 78.4 \\
\midrule
EigenScore-B & 74.2 & 68.5 & 72.7 & 75.0 & 79.0 & 73.4 & 63.7 & 84.0 \\
\rowcolor{lightgray}
EigenScore-B+DMP & 71.3 & 65.8 & 73.7 & 75.1 & 78.9 & 71.4 & 65.1 & 81.8 \\
\midrule
\midrule

\textbf{Mean} -- Baseline & 74.09 & \textbf{69.53} & 73.54 & 69.73 & 77.70 & 71.17 & \textbf{67.49} & \textbf{79.69} \\
\rowcolor{lightgray}
\textbf{Mean} -- DMP \textbf{(Ours)} & \textbf{74.11} & 69.04 & \textbf{74.01} & \textbf{71.03} & \textbf{77.79} & \textbf{71.31} & 67.30 & 79.68 \\
\textbf{Time (s)} -- Baseline & 1778 & 2368 & 1932 & 2148 & 1124 & 1750 & 1430 & 1225 \\
\rowcolor{lightgray}
\textbf{Time (s)} -- DMP \textbf{(Ours)} & 627 (2.8x) & 1246 (1.9x) & 1022 (1.9x) & 904 (2.4x) & 442 (2.5x) & 927 (1.9x) & 744 (1.9x) & 630 (2.0x)   \\
\textbf{Reuse Ratio} & 66.8\% & 52.7\% & 50.4\% & 62.6\% & 63.4\% & 51.3\% & 51.3\% & 55.1\% \\
\bottomrule
\end{tabular}
\caption{AUROC, mean AUROC, running time, and reuse ratio of different baselines using LLaMA-2-7B-Chat and Mistral-7B-Instruct on TriviaQA, NQ Open, SQuAD, and HaluEval. The best mean AUROC is highlighted in bold.}
\label{tab:main-table}
\end{table*}

\begin{remark}
   Although our primary focus in this paper is on hallucination detection, the applicability of DMP extends beyond this use case. Self-consistency methods are broadly employed in tasks such as uncertainty quantification~\citep{kuhn2023semantic}, alignment~\citep{amini2024variational}, and reasoning~\citep{zuo2025ttrl} in LLMs. 
\end{remark}

\section{Experiments}
\label{sec:experiments}

In this section, we first describe our experimental settings. We then evaluate the performance of our proposed DMP across multiple datasets using self-consistency methods, demonstrating its ability to accelerate inference without compromising accuracy. Furthermore, we conduct comprehensive ablation studies to analyze the contributions of selective inference and annealed decoding.

\subsection{Experiment Settings}
\label{subsec:experiemt_settings}
\textbf{Datasets.} We evaluate the effectiveness of our proposed DMP on four generative question-answering (QA) datasets: TriviaQA~\citep{joshi2017triviaqa}, NQ Open~\citep{kwiatkowski2019natural}, SQuAD~\citep{rajpurkar2016squad}, and the QA subset of HaluEval~\citep{li2023halueval}. To further demonstrate the robustness of our method, we include additional results on SciQ~\citep{welbl2017crowdsourcing} in Appendix E, showcasing its performance on domains requiring specialized knowledge. For each dataset, we use 400 test examples randomly sampled from the original larger dataset.

\textbf{Models.} Following prior work on hallucination detection, we conduct experiments using widely adopted open-source LLMs, including LLaMA-2-7B-Chat~\citep{touvron2023llama} and Mistral-7B-Instruct~\citep{jiang2023clip}. To demonstrate the scalability and generalization of our DMP, we present additional experiments on LLaMA-2-13B-Chat and Falcon-7B-Instruct~\citep{penedo2023refinedweb}, provided in Appendix D.2.

\textbf{Evaluation.} Following~\citet{farquhar2024detecting}, we use GPT-4o~\citep{hurst2024gpt} to label the correctness of generated responses, as standard metrics often fall short for sentence-level evaluation. We assess detection performance using the area under the receiver operating characteristic curve (\textbf{AUROC}). To quantify the generation cost, we report the average generation time per input prompt on a single H100 GPU. To factor out hardware and system effects, we also report the reuse ratio, defined as the percentage of generated tokens for which the forward process is skipped, thereby reflecting the theoretical speedup.

\textbf{Baselines.} We evaluate our proposed DMP against several widely-used self-consistency baselines, including LN-Entropy\citep{malinin2020uncertainty}, Lexical Similarity\citep{lin2022towards}, Semantic Entropy\citep{farquhar2024detecting}, SelfCheckGPT\citep{manakul2023selfcheckgpt}, and EigenScore~\citep{chen2024inside}. Moreover, we provide additional experiments on the black box version of Semantic Entropy and EigenScore as``Semantic Entropy-B'' and ``EigenScore-B''. For detailed descriptions of these baselines, please refer to Appendix A. 

\textbf{Implementation.} We implement our selective inference using the KV Cache variant. 
We adopt the following default settings. We set the generation temperature to 0.8. For annealed decoding, we set the importance-selection threshold $\alpha$ to $0.9$ and the annealing speed $\eta$ to $1.4$. For hard decoding, we set the confidence threshold $\gamma$ to $0.8$. 
For additional implementation details, 
please refer to Appendix B.2. For notational simplicity, we denote our method built upon a baseline implementation as ``\textbf{+DMP}''. For example, when applied to LN-Entropy, we denote it as ``\textbf{LN-Entropy+DMP}''.

\subsection{Main Results}
\label{subsec:main_results} 
Table~\ref{tab:main-table} reports the AUROC, mean AUROC, running time, and reuse ratio across different models and datasets, with the best mean performance highlighted in bold. Based on these results, we draw the following conclusions.

\textbf{Detection Performance.} Our method consistently maintains a high AUROC across baselines, models, and datasets, in some cases even slightly exceeding baseline performance. Specifically, the AUROC reduction is at most $2.9\%$ across all trials, while the mean AUROC reduction over all self-consistency methods is limited to just $0.5\%$, both well within an acceptable range. Notably, the mean AUROC with our DMP surpasses the baseline in 5 out of 8 trials, indicating that the introduced sampling randomness does not compromise, and may even enhance the detection performance.

\textbf{Efficiency Improvement.} Applying DMP consistently improves the efficiency of self-consistency methods, as confirmed by both theoretical analysis and empirical results. In terms of reuse ratio, our approach skips up to $66.8\%$ of token generation in TriviaQA with Llama2-7B-Chat, achieving a $3\times$ acceleration. The measured running time closely matches the theoretical speedup on hardware. While the gains are smaller on more challenging datasets such as NQ Open and SQuAD, our method still consistently delivers at least a $2\times$ acceleration in both theory and $1.9\times$ practice.

\subsection{Ablation Study}
\label{subsec:ablation_study}
In this section, we present comprehensive ablation studies to examine the contributions of individual components of DMP, reporting mean measurements over the baselines. We first analyze the effect of the hard decoding threshold, followed by evaluating the effectiveness of annealed decoding. Furthermore, we assess the effect of sampling temperature on acceleration performance in Appendix D.3 and provide memory analysis in Appendix F. We denote w/ SI as the method that applies selective inference, and w/ Hard denotes the method applying hard decoding. We further denote Baseline as standard generation. The small error bar indicates the reuse ratio of the corresponding setting.

\begin{figure}[!t]
    \centering
    \begin{subfigure}{0.49\linewidth}
        \centering
        \includegraphics[width=\linewidth]{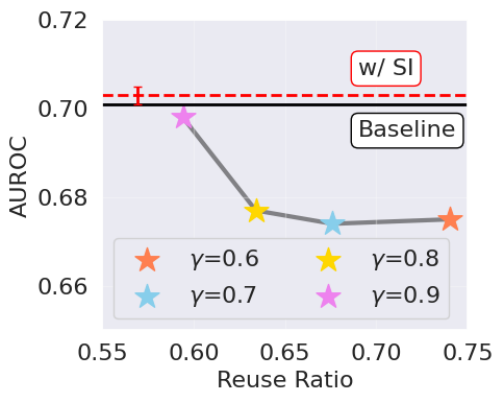}
        \caption{TriviaQA}
        \label{fig:hard_decoding_triviaqa}
    \end{subfigure}
    \hfill
    \begin{subfigure}{0.49\linewidth}
        \centering
        \includegraphics[width=\linewidth]{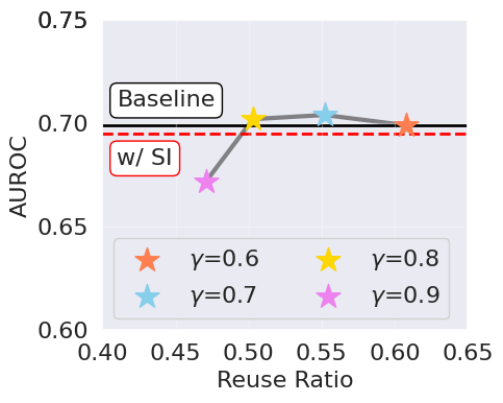}
        \caption{SQuAD}
        \label{fig:hard_decoding_squad}
    \end{subfigure}
    
    \caption{AUROC and reuse ratio of confidence threshold study in hard decoding. Baseline denotes standard generation, while w/ SI refers to selective inference without other techniques. The error bar indicates the reuse ratio of w/ SI.}
    \label{fig:hard_decoding}
\end{figure}

\begin{figure}[!t]
    \centering
    \begin{subfigure}{0.49\linewidth}
        \centering
        \includegraphics[width=\linewidth]{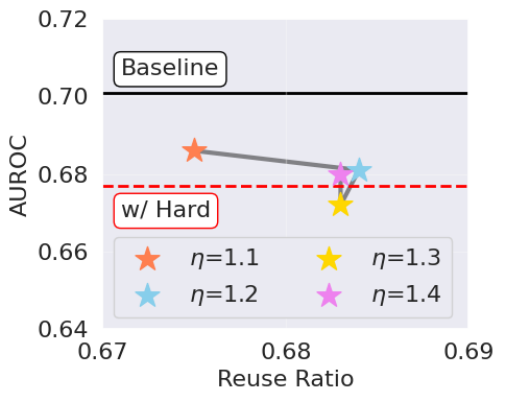}
        \caption{$\gamma=0.75$}
        \label{fig:rescaling_075}
    \end{subfigure}
    \hfill
    \begin{subfigure}{0.49\linewidth}
        \centering
        \includegraphics[width=\linewidth]{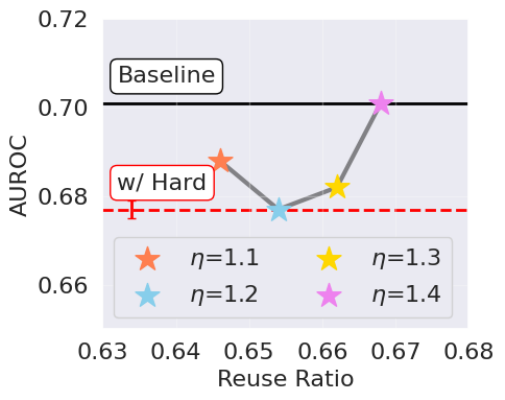}
        \caption{$\gamma=0.8$}
        \label{fig:rescaling_08}
    \end{subfigure}
    
    \caption{AUROC and reuse ratio of rescaling study in annealed decoding. Baseline denotes standard generation, while w/ Hard refers to hard decoding. The red error bar indicates the reuse ratio of w/ Hard.}
    \label{fig:rescaling}
\end{figure}

\textbf{Effectiveness of Hard Decoding.} We conduct experiments on TriviaQA and SQuAD using Llama2-7B-Chat, examining the trade-off between AUROC and reuse ratio under different confidence thresholds $\gamma \in \{0.6, 0.7, 0.8, 0.9\}$. As shown in Figure~\ref{fig:hard_decoding}, AUROC remains robust across different values of $\gamma$, while the reuse ratio increases as $\gamma$ decreases. We conservatively select $\gamma=0.8$ as the final threshold, as it provides a favorable balance. Notably, hard decoding does not necessarily reduce performance, suggesting that existing self-consistency methods may be underconfident.

\textbf{Effectiveness of Annealed Decoding.} We conduct experiments on the rescaling hyperparameter $\eta \in \{1.1, 1.2, 1.3, 1.4\}$ in annealed decoding to study its impact on TriviaQA using Llama2-7B-Chat under two confidence thresholds $\gamma$. As shown in Figure~\ref{fig:rescaling}, annealed decoding yields improvements over hard decoding and can even approach baseline performance on both efficiency and detection performance. We attribute this to the rescaling, which makes non-exact-answer tokens more deterministic, while hard decoding further enhances consistency and mitigates underconfidence in non-exact-answer tokens.

\begin{figure}[!t]
    \centering
    \begin{subfigure}{0.49\linewidth}
        \centering
        \includegraphics[width=\linewidth]{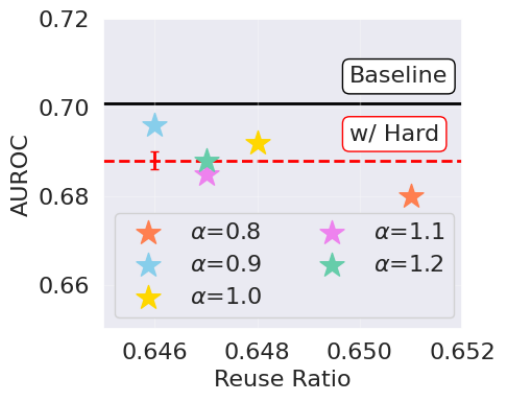}
        \caption{$\eta=1.1$}
        \label{fig:threshold11}
    \end{subfigure}
    \hfill
    \begin{subfigure}{0.49\linewidth}
        \centering
        \includegraphics[width=\linewidth]{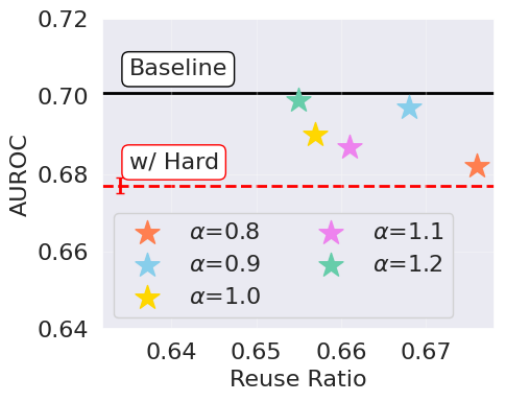}
        \caption{$\eta=1.4$}
        \label{fig:threshold14}
    \end{subfigure}
    
    \caption{AUROC and reuse ratio of selection threshold study in annealed decoding. Baseline denotes standard generation, while w/ Hard refers to hard decoding. The red error bar indicates the reuse ratio of w/ Hard.}
    \label{fig:threshold}
\end{figure}

Additionally, we conduct experiments on the selection threshold $\alpha \in \{0.8, 0.9, 1.0, 1.1, 1.2\}$ in annealed decoding to study its impact on TriviaQA using Llama2-7B-Chat under two rescaling parameters. As shown in Figure~\ref{fig:threshold}, setting $\alpha = 1.2$ consistently maintains a high AUROC while achieving a favorable trade-off. Finally, we present performance improvement applying our techniques in Table~\ref{tab:component}.

\begin{table}[!t]
\centering
\small
\begin{tabular}{l|c|c|c|c}
\toprule
& Baseline & \makecell{+Selective \\ Inference} & \makecell{+Hard \\ Decoding} & \makecell{+Annealed \\ Decoding} \\
\midrule
AUROC & 70.1 & 70.2 & 67.7 & 69.7 \\
Reuse Ratio & -- & 56.9\% & 63.4\% & 66.8\% \\
\bottomrule
\end{tabular}
\caption{AUROC and reuse ratio for technique components.}
\label{tab:component}
\end{table}

\section{Conclusion}
\label{sec:conclusion}
In this work, we present the Decoding Memory Pipeline (DMP), a novel approach to accelerate hallucination detection in large language models by leveraging redundancy in self-consistency decoding. By identifying and reusing repeated tokens across generations, DMP significantly reduces inference cost while maintaining strong detection performance. Our experiments demonstrate up to a 3x speedup without compromising AUROC scores, making DMP practical for real-world applications. A limitation of our approach is the increased memory cost resulting from caching more responses than standard generation. Developing efficient cache storage strategies is left for future work.

\section{Acknowledgment}
This research was supported by the Graduate Research at ORNL (GRO) Internship Program at Oak Ridge National Laboratory and used resources of the Oak Ridge Leadership Computing Facility (OLCF), which is a DOE Office of Science User Facility at the Oak Ridge National Laboratory supported by the U.S. Department of Energy under Contract No. DE-AC05-00OR22725.

\bibliography{aaai2026}

\newpage
\appendix

\section{Background}
\label{app:background}
In this section, we present the background knowledge about KV cache and the self-consistency baselines used in our paper.

\textbf{KV cache.} To reduce the computational burden of the attention mechanism, past key and value tensors are cached to avoid redundant computation. During generation, the past tokens involve the logit computation of the next token $y_{i+1}$ via the attention: 
\begin{align}
    & \text{Attn}_i = \text{softmax}\left(\frac{\vQ_i \vK_{1:i}^T}{\sqrt{d_k}}\right)\vV_{1:i},\\
    & \vQ_{i} = \vX_{i}\vW^Q, \vK_{1:i} = \vX_{1:i}\vW^K, \vV_{1:i} = \vX_{1:i}\vW^V, 
\end{align}
where $d_K$ denotes the dimensionality of the keys, $\vX_{1:i}$ denotes the input of attention, and $\vW^Q$, $\vW^K$, and $\vW^V$ represent the linear projections for the query, key, and value, respectively. Due to the causal mask in the attention mechanism, $\vK_{1:i-1}$ and $\vV_{1:i-1}$ remain fixed during generation. Thus, only $K_i$ and $V_i$ need to be computed at each step, followed by concatenation with $\vK_{1:i-1}$ and $\vV_{1:i-1}$. Consequently, we can update Equation 1 in the main paper as follows:
\begin{align}
    \vs_i = f_\theta(\vx, y_{i-1}, \vK_{1:i-2}, \vV_{1:i-2}), \quad i = 2, 3, \ldots, L.
\end{align}
Despite the optimizations introduced by KV cache, the sequential nature of generation remains computationally intensive, as it requires $O(L)$ model forward passes for a sequence of length $L$.

\textbf{LN-Entropy.} Given $N$ responses $\cY=\{\vy_1, \vy_2, \dots, \vy_N\}$ and the input prompt $\vx$, length-normalized entropy computes the consistency score as follows:
\begin{align}
    S_\text{LNE}(\cY|\vx) = -\mathbb{E}_{\vy\in\cY}\frac{1}{L_\vy}\sum_{i=1}^{L_\vy}\log\vP(\vy_i \mid \vx, \vy_{1:i-1}), 
\end{align}
where $L_\vy$ denotes the length of the response $\vy$. LN-Entropy computes the consistency score as the average log-likelihood of the responses, normalized by response length.

\textbf{Lexical Similarity.} Given $N$ responses $\cY=\{\vy_1,\vy_2, \dots, $ $ \vy_N\}$ and the input prompt $\vx$, lexical similarity estimates the consistency score by constructing semantic graphs and measuring similarities based on the uncertainty of a sequence-to-sequence model as follows:
\begin{align}
    S_\text{LEX}(\cY|\vx) = \frac{1}{N(N-1)}\sum_{i=1}^N\sum_{j=i+1}^N \textit{sim}(\vy_i, \vy_j),
\end{align}
where \textit{sim}$(\cdot, \cdot)$ is the similarity function using semantic rules. We compute lexical similarity using the ROUGE metric.\footnote{\url{https://huggingface.co/spaces/evaluate-metric/rouge}}

\textbf{SelfCheckGPT.} Given $N$ responses $\cY=\{\vy_1,\vy_2, \dots, $ $ \vy_N\}$ and the input prompt $\vx$, SelfCheckGPT evaluates the consistency between generations using methods such as BERTScore, designed QA, NLI tasks, or n-gram models. In our work, we adopt BERTScore, which is computed as follows:
\begin{align}
    S_\text{BERT}(\cY|\vx) = 1-\frac{1}{N}\sum_{i=1}^N \cB(\vy_i, \hat\vy),
\end{align}
where $\cB$$(\cdot, \cdot)$ represents the BERTScore computation function, and $\hat\vy$ is the most probable answer generated with greedy decoding. We choose NLI-RoBERTa-Large as the backbone of the score computation\footnote{\url{https://huggingface.co/sentence-transformers/nli-roberta-large}}.

\textbf{EigenScore.} Given $N$ responses $\cY=\{\vy_1,\vy_2, \dots, $ $ \vy_N\}$ and the input prompt $\vx$, EigenScore measures the consistency between answers with the eigenvalues of the sentence embedding matrix. To be specific, the sentence embedding matrix is the stack of the hidden states of the last token:
\begin{align}
    \vZ = \textit{Concat}([\vh_{1}[-1], \vh_{2}[-1], \dots, \vh_{N}[-1]]),
\end{align}
where \textit{Concat}$(\cdot)$ denotes concatenation along the first dimension, and $\vh_{i}[-1]$ refers to the hidden state of the last token in response $i$. The covariance matrix of $\vZ$ is then computed as:
\begin{align}
    \Sigma = \vZ^\top (\mathbf{I}-\frac{1}{N}\mathbf{1}^\top\mathbf{1})\vZ,
\end{align}
where $\mathbf{1}$ denotes an all-ones column vector. The EigenScore is then computed using the determinant of $\Sigma$:
\begin{align}
    S_\text{Eigen} = \frac{1}{N}\sum_{i=1}^N\log\det(\Sigma+\alpha\mathbf{I}),
\end{align}
where $\alpha$ is a regularization term introduced to mitigate issues arising from singular values. 

In black-box scenarios, where the hidden states of LLMs are inaccessible, we employ a proxy model to compute the hidden states. We denote this variant as \textbf{EigenScore-B}, using NLI-RoBERTa-Large$^2$ as the proxy model.

\textbf{Semantic Entropy.} Given $N$ responses $\cY=\{\vy_1,\vy_2, \dots, $ $ \vy_N\}$ and the input prompt $\vx$, semantic entropy first clusters the responses in to clusters $\cC$ with the help of other language models. Then the consistency score is computed through a Rao-Blackwellized Monte Carlo integration over the semantic
equivalence classes $\cC$:
\begin{align}
    S_\text{SE}(\cY|\vx) = -\sum_{c\in\cC}\vP(c|\vx)\log \vP(c|\vx), \\
    P(c|\vx) = \sum_{\vy_i\in c} \vP(\vy_i|\vx)
\end{align}
where $P(c|\vx)$ is normalized over the sampled clusters. We use GPT-3.5-Turbo to generate cluster responses. For the specific prompt, please refer to Appendix B.2. 

In black-box scenarios, where the sentence likelihood of LLMs is inaccessible for the Rao-Blackwellized formulation, we estimate semantic entropy by counting the samples in each cluster. We denote this variant as \textbf{Semantic Entropy-B}.

\section{Implementation Details}
In this section, we present the implementation details of the implementation of our preliminary study and main experiments.

\subsection{Preliminary Study}
\label{app:implement_prelim}
We show how the prefix sharing proportion is computed and how we construct the prompt for splitting the answers into two groups.

\textbf{Prefix Sharing Proportion.} To compute the prefix shared proportion of Observation 1, we use the sentence level and word level, respectively. Given $M$ input prompts $\vx^{(i)}$, we generate $N$ responses $\vy_j^{(i)}$, where $i=1,2,\dots,M$, and $j=1,2,\dots,N$. The sentence-level prefix sharing proportion is computed as follows:
\begin{align}
    p_\text{sentence} =& \frac{2}{MN(N-1)} \\
    &*\sum_{m=1}^M\sum_{i=1}^N\sum_{j=i+1}^N\mathbf{1}\{\textit{Entail}(\vy_i^{(m)}, \vy_j^{(m)})\},
\end{align}
where $\mathbf{1}$ is the indicator function and \textit{Entail($\vy_i^{(m)}, \vy_j^{(m)}$)} represents either $\vy_i^{(m)}$ starts with $\vy_j^{(m)}$ or $\vy_j^{(m)}$ starts with $\vy_i^{(m)}$. The word-level prefix sharing proportion is computed as:
\begin{align}
    p_\text{word} =& \frac{2}{MN(N-1)} \\
    &*\sum_{m=1}^M\sum_{i=1}^N\sum_{j=i+1}^N\mathbf{1}\{\textit{Entail}(\vy_i^{(m)}, \vy_j^{(m)})\} \\
    &*\frac{\min(L_{\vy_i^{(m)}}, L_{\vy_j^{(m)}}}{\sum_{m=1}^M\sum_{i=1}^NL_{\vy_i^{(m)}}},
\end{align}
where $L_{\vy_i^{(m)}}$ represents the length of the response $\vy_i^{(m)}$. For each model and dataset, we generate 10 responses for 400 questions using a sampling temperature of $0.8$.

\textbf{Grouping Prompt} For Observation 2, we group responses into two groups according to the answer template. We use GPT-3.5-Turbo for grouping with the following prompt:

\textit{``Given the question, please split the given responses into two groups according to the structure or template of the response. Please return two index list for the two groups.}

\textit{Question: \{Question\}}

\textit{Response 1: \{Response 1\}}

\textit{Response 2: \{Response 2\}}

\textit{\vdots }

\textit{Response 10: \{Response 10\}}

\textit{Returned Lists: ''}

In our experiments, we sample 100 questions for each dataset and generate 10 responses for each question. To label the correctness of the response, we use GPT-3.5-Turbo and use the same labeling method as the main experiments (Appendix B.2).

\subsection{Main Experiments}
\label{app:implement_main}
In this section, we present the implementation details of our main experiments, including the prompt design for sentence-level response generation, the prompt design for response labeling, the short-answer cut technique in annealed decoding, and the selection of specific hyperparameters.

\textbf{Prompt for Sentence Generation.} Following the approach of semantic entropy~\citep{farquhar2024detecting}, we use the following prompt to generate sentence-level responses:

\textit{Answer the following question in a single brief but complete
sentence.}

\textit{Question: \{question\}}

\textit{Answer:}

\textbf{Prompt for Response Labeling.} Following the approach of semantic entropy~\citep{farquhar2024detecting}, we use the following prompt for response labeling when the dataset contains only a single reference answer:

\textit{We are assessing the quality of answers to the following question: \{question\}}

\textit{The expected answer is: \{reference answer\}}

\textit{The proposed answer is: \{predicted answer\}}

\textit{Within the context of the question, does the proposed answer mean the same as the expected answer? Respond only with yes or no.}

For datasets with multiple reference answers, we employ the following prompt with small modifications:

\textit{We are assessing the quality of answers to the following question: \{question\}}

\textit{The following are expected answers to this
question:}

\textit{The proposed answer is: \{predicted answer\}}

\textit{Within the context of the question, does the proposed answer mean the same as any of the expected answers? Respond only with yes or no.}

\textbf{Prompt for Entailment Estimating.} In semantic entropy, we need to cluster the responses. We use the following prompt with GPT-3.5-Turbo:

\textit{We are evaluating answers to the question \{question\}}

\textit{Here are two possible answers:}

\textit{Possible Answer 1: \{text1\}}

\textit{Possible Answer 2: \{text2\}}

\textit{Does Possible Answer 1 semantically entail Possible Answer 2?}

\textit{Respond with entailment, contradiction, or neutral.}

\textbf{Dealing with Short Answer in Annealed Decoding.} Our proposed annealed decoding method searches for non-exact-answer tokens. However, in some cases, the generated responses are already exact answers under our generation prompt. To avoid unnecessary errors, we identify such short answers and exclude them from annealed decoding. In our experiments, we define short answers as responses containing fewer than 10 tokens.

\textbf{SQuAD Experimental Settings.} For SQuAD, we find that experiments with annealed decoding perform significantly worse than those using only selective inference with hard decoding, which also provides greater efficiency in generation. Therefore, we report hard decoding results for our experiments on SQuAD. We attribute this outcome to the inherent difficulty of the dataset.

\section{Batch-Wise Algorithm}
\label{app:algorithm}
In this section, we present the algorithm for batch-wise selective inference in Algorithm~\ref{alg:batch}. In the algorithm, \textit{Update}($\cdot$) denotes the token-level information update in selective inference, while \textit{Next}($\cM$) retrieves the next cached response from $\cM$. Note that although all update rules during generation are applied in parallel with the mask, we present them in an instance-wise manner for improved readability.

\begin{algorithm}[!h]
\small
\caption{Batch-Wise Decoding Memory Pipeline}
\textbf{Input}: Input Batch $\vx$, Number of Responses $N$, model $f$\\
\textbf{Parameter}: Sampling Temperature $T$, Confidence Threshold $\gamma$, Rescaling $\eta$, Selection Threshold $\alpha$\\
\textbf{Output}: Batched $N$ responses
\begin{algorithmic}[1] %[1] enables line numbers
\STATE Initialize memory list $\cM_{j}=[\quad]$ for $j$-th instance in the batch;
\STATE Initialize the reference batch $B_R=\textit{None}$
\FOR{$i=1:N$}
\STATE Initialize the attention mask $\mu_a$;
\STATE Initialize the reusing mask $\mu_r=\mu_a$;
\STATE Initialize the current generation batch $B_{G_i}=\vx$;
\FOR{$j=1:$max\_generation} 
\item\textit{\# Update the following in parallel with $\mu_r$}
\IF{$\mu_a(B_{G_i})[j]==\mu_a(B_R)[j]$ is True for at least one $j$}
\STATE $\mu_r(B_{G_i}) = \textit{Update}(\mu_r(B_R))$;
\STATE $\mu_r(\mu_a) = \mu_r$;
\STATE $B_R[j] = \textit{Next}(\cM[j])$ for $\mu_a(B_{G_i}) $ $[j]\neq\mu_a(B_R)[j]$;
\ELSE
\STATE $B_{G_i}, \mu_a$ = $f(B_{G_i}, \mu_a, K_i, V_i)$;
\ENDIF
\ENDFOR
\ENDFOR
\IF {$B_{G_i}[j]$ is not contained in $\cM_j$}
\STATE $\cM_j$.append(($B_{G_i}[j]$, $\vs_i$, $K_i$, $V_i$, $\vh_i$));
\ELSE
\STATE Anneal the memory in $\cM$ corresponding to $B_{G_i}[j]$;
\ENDIF
\STATE \textbf{return} $[B_{G_1}, B_{G_2}, \dots, B_{G_N}]$
\end{algorithmic}
\label{alg:batch}
\end{algorithm}

\section{Additional Results}
In this section, we present additional experiments, including preliminary studies, main experiments on other models, and an ablation study on sampling temperature selection. Unless otherwise noted, the experimental settings are consistent with those used in the main experiments.

\subsection{Preliminary Study on Other Models}
\label{app:results_models_prelim}
\begin{figure}[!th]
    \centering
    \includegraphics[width=0.7\linewidth]{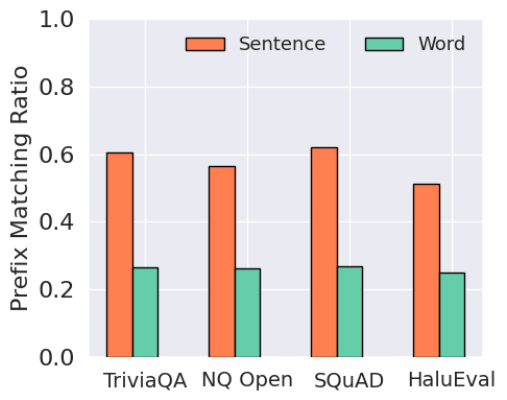}
    \caption{Prefix matching ratios at the sentence level and word level across datasets with Mistral-7B-Instruct.}
    \label{fig:app_prefix_mistral}
\end{figure}

\begin{figure}[!th]
    \centering
    \includegraphics[width=0.7\linewidth]{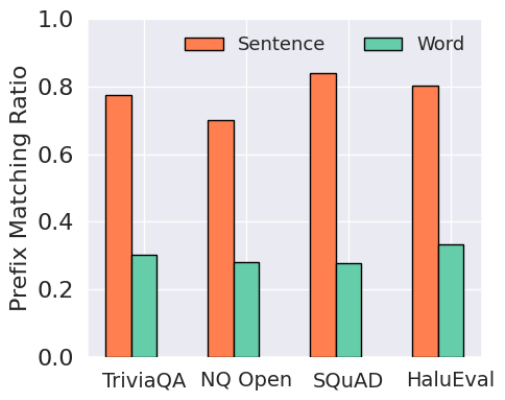}
    \caption{Prefix matching ratios at the sentence level and word level across datasets with Llama2-13B-Chat.}
    \label{fig:app_prefix_llama2-13B}
\end{figure}

\begin{figure}[!th]
    \centering
    \includegraphics[width=0.7\linewidth]{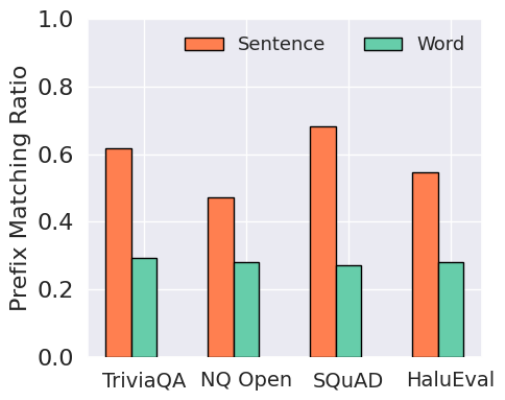}
    \caption{Prefix matching ratios at the sentence level and word level across datasets with Falcon-7B-Instruct.}
    \label{fig:app_prefix_falcon}
\end{figure}

As shown in the main paper, sentence-level and word-level prefix sharing frequently occurs in Llama2-7B-Chat across datasets. Here, we provide statistical results for other models in Figures~\ref{fig:app_prefix_mistral}, \ref{fig:app_prefix_llama2-13B}, and \ref{fig:app_prefix_falcon}. The results indicate that redundancy is prevalent across different models.

\subsection{Main Experiments on Other Models}
\label{app:results_models_main}
We report experiments on LLaMA2-13B-Chat and Falcon-7B-Instruct to demonstrate the scalability and generalization of our DMP. To reduce the cost of OpenAI API usage, we omit the semantic entropy results. As shown in Table~\ref{tab:app_main_table}, our method consistently improves generation efficiency while maintaining high AUROC across both scalable models and others.

\begin{table*}[!ht]
\centering
\small
\begin{tabular}{l|cccc|cccc}
\toprule
\multicolumn{1}{c|}{} & 
\multicolumn{4}{c|}{\textbf{Llama2-13B-Chat}} & 
\multicolumn{4}{c}{\textbf{Falcon-7B-Instruct}} \\
\multicolumn{1}{c|}{} & \textbf{TriviaQA} & \textbf{NQ Open} & \textbf{SQuAD} & \textbf{HaluEval} & \textbf{TriviaQA} & \textbf{NQ Open} & \textbf{SQuAD} & \textbf{HaluEval} \\
\midrule
LN-Entopy & 0.686 & 0.594 & 0.697 & 0.666 & 0.608 & 0.603 & 0.507 & 0.659 \\
\rowcolor{lightgray}
LN-Entropy+DMP & 0.713 & 0.636 & 0.736 & 0.686 & 0.621 & 0.638 & 0.573 & 0.689 \\
\midrule
Lexical Similarity & 0.652 & 0.588 & 0.730 & 0.574 & 0.689 & 0.628 & 0.627 & 0.767 \\
\rowcolor{lightgray}
Lexical Similarity+DMP & 0.646 & 0.599 & 0.727 & 0.599 & 0.692 & 0.618 & 0.623 & 0.746 \\
\midrule
SelfCheckGPT & 0.633 & 0.623 & 0.753 & 0.605 & 0.651 & 0.597 & 0.623 & 0.732 \\
\rowcolor{lightgray}
SelfCheckGPT+DMP & 0.641 & 0.632 & 0.742 & 0.624 & 0.672 & 0.616 & 0.642 & 0.693 \\
\midrule
EigenScore & 0.701 & 0.625 & 0.776 & 0.657 & 0.717 & 0.642 & 0.703 & 0.786 \\
\rowcolor{lightgray}
EigenScore+DMP & 0.702 & 0.628 & 0.754 & 0.660 & 0.710 & 0.630 & 0.647 & 0.768 \\
\midrule
EigenScore-B & 0.725 & 0.647 & 0.754 & 0.701 & 0.725 & 0.666 & 0.728 & 0.799 \\
\rowcolor{lightgray}
EigenScore-B+DMP & 0.718 & 0.641 & 0.737 & 0.698 & 0.724 & 0.664 & 0.690 & 0.776 \\
\midrule
\midrule

\textbf{Mean} -- Baseline & 0.679 & 0.615 & \textbf{0.742} & 0.641 & 0.678 & 0.627 & \textbf{0.637} & \textbf{0.749} \\
\rowcolor{lightgray}
\textbf{Mean} -- DMP \textbf{(Ours)} & \textbf{0.684} & \textbf{0.627} & 0.739 & \textbf{0.653} & \textbf{0.684} & \textbf{0.633} & 0.635 & 0.734 \\
\textbf{Reuse Ratio} & 0.679 & 0.515 & 0.499 & 0.636 & 0.639 & 0.521 & 0.523 & 0.594 \\
\bottomrule
\end{tabular}
\caption{AUROC, mean AUROC, running time, and reuse ratio of different baselines using LLaMA-2-13B-Chat and Falcon-7B-Instruct on TriviaQA, NQ Open, SQuAD, and HaluEval. The best mean AUROC is highlighted in bold.}
\label{tab:app_main_table}
\end{table*}

\begin{figure}[!th]
    \centering
    \includegraphics[width=0.7\linewidth]{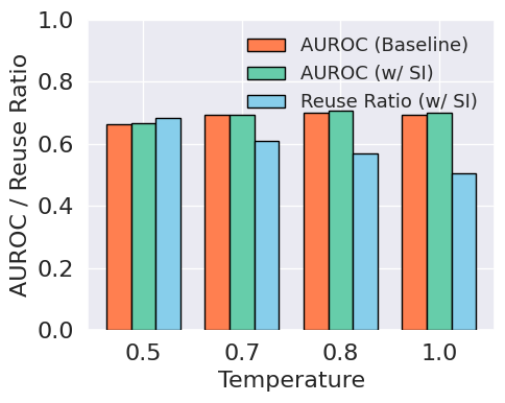}
    \caption{AUROC and reuse ratio of selective inference compared to the baseline under different sampling temperatures on TriviaQA. w/ SI represents selective inference.}
    \label{fig:temperature_triviaqa}
\end{figure}

\begin{figure}[!th]
    \centering
    \includegraphics[width=0.7\linewidth]{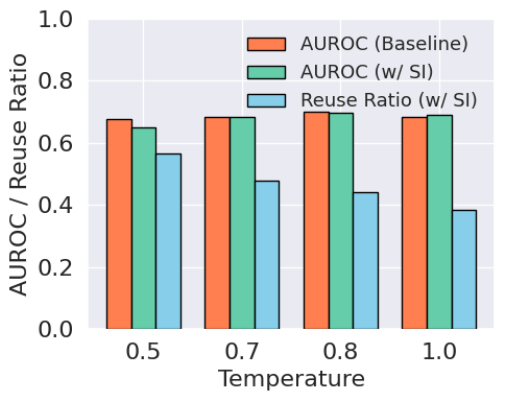}
    \caption{AUROC and reuse ratio of selective inference compared to the baseline under different sampling temperatures on SQuAD. w/ SI represents selective inference.}
    \label{fig:temperature_squad}
\end{figure}

\subsection{Ablation Study of Sampling Temperature}
\label{app:ablation_temperature}
Sampling temperature can influence the performance of hallucination detection. A higher temperature is often desirable for capturing uncertainty, while our method instead leverages redundancy, where a lower temperature increases the proportion of response reuse. To study this trade-off, we conduct experiments with $T \in \{0.5, 0.7, 0.8, 1.0\}$ on both the baseline and selective inference (denoted as w/ SI). As shown in Figure~\ref{fig:temperature_triviaqa} and~\ref{fig:temperature_squad}, selective inference maintains generation quality while substantially improving efficiency. For a conservative balance, we adopt $T=0.8$ as the final sampling temperature.

\begin{table}[!t]
\centering
\small
\begin{tabular}{l|cc}
\toprule
& \makecell{Llama2- \\ 7B-Chat} & \makecell{Mistral- \\ 7B-Instruct} \\
\midrule
LN-Entopy  & 62.6 & 62.6 \\
\rowcolor{lightgray}
LN-Entropy+DMP & 66.5 & 60.3 \\
\midrule
Lexical Similarity & 63.3 & 63.3 \\
\rowcolor{lightgray}
Lexical Similarity+DMP & 65.1 & 68.6 \\
\midrule
EigenScore & 68.4 & 68.4 \\
\rowcolor{lightgray}
EigenScore+DMP & 67.2 & 69.8 \\
\midrule
EigenScore-B & 67.7 & 67.7 \\
\rowcolor{lightgray}
EigenScore-B+DMP & 66.6 & 72.1 \\
\midrule
\midrule

\textbf{Mean} -- Baseline & 65.5 & 65.5 \\
\rowcolor{lightgray}
\textbf{Mean} -- DMP \textbf{(Ours)} & \textbf{66.35} & \textbf{67.7} \\
\rowcolor{lightgray}
\textbf{Reuse Ratio} & 65.5\% & 57.1\% \\
\bottomrule
\end{tabular}
\caption{AUROC, mean AUROC, running time, and reuse ratio of different baselines using LLaMA-2-7B-Chat and Mistral-7B-Instruct on SciQ. The best mean AUROC is highlighted in bold.}
\label{tab:app-sciq}
\end{table}

\section{Results on SciQ}
\label{app:results_sciq}
To verify that our method is not domain-dependent, which is a key advantage of self-consistency methods, we conduct experiments on SciQ~\citep{welbl2017crowdsourcing} using LLaMA2-7B-Chat and Mistral-7B-Instruct. To reduce time and cost, we evaluate only four baselines: LN-Entropy, Lexical Similarity, EigenScore, and EigenScore-B. As shown in Table~\ref{app:results_sciq}, our DMP achieves significant efficiency gains on SciQ, and the mean AUROC even surpasses that of the baselines.

\section{Memory Cost}
\label{app:memory}
One limitation of our DMP is that it requires storing more cache compared to standard generation. To quantify the additional memory cost, we conduct experiments with LLaMA2-7B-Chat on TriviaQA using a batch size of 10. The baseline exhibits a peak reserved memory of 28,312 MB, whereas our DMP requires 29,034 MB, with an increase of 722MB.

\end{document}